\documentclass{llncs}
\usepackage{times}
\usepackage{helvet}
\usepackage{courier}
\usepackage{makeidx}
\usepackage{amssymb,latexsym,amsmath}     
\usepackage{graphicx,array}
\usepackage{enumitem}
\usepackage{booktabs}
\usepackage{multirow} 

\usepackage{multicol} 
\usepackage{float}
\usepackage{algorithm}
\usepackage{algorithmic}
\usepackage[table,x11names]{xcolor}
\usepackage{pgfplots}
\usepackage{caption}
\usepackage[autostyle]{csquotes}
\usepackage{commath}
\usepackage{hhline}
\usepackage{soul}
\usepackage{tikz}
\usepackage{csquotes}
\usepackage{rotating}
\usetikzlibrary{patterns}

\renewcommand\vec[1]{\overrightarrow{#1}}
\newcommand\cev[1]{\overleftarrow{#1}}
\def\,{\nobreak\hspace{.16667em plus .08333em}}

\def\qbegin{
\begin{tabular}{ @{}c@{~~} l@{\qquad}  >{\tt}r@{} }
 \midrule
 \multicolumn{2}{l}{Choice set}&{$Closeness$}\\ 
 \midrule
}
\def\qend#1{
 \midrule
 \multicolumn{2}{r}{\emph{Difficulty}} & #1\\
 \midrule
\end{tabular}
}

\def\qbeginnew{
\begin{tabular}{ @{}ll@{}}
 \midrule
 \multicolumn{2}{l}{Choice set}\\ 
 \midrule
}

\begin{document}

\title{ Modeling of Item-Difficulty for Ontology-based Multiple-Choice Questions}
\titlerunning{Item-Difficulty} 
\author{Vinu E.V\inst{1},  Tahani Alsubait\inst{2}
\and P Sreenivasa Kumar\inst{1}}

\institute{Department of Computer Science and Engineering, Indian Institute of Technology Madras, Chennai, India \and Information Science Department, Umm Alqura University, Makkah, Saudi Arabia\\
\email{\{vinuev, psk\}@cse.iitm.ac.in},\\ 
\email {tmsubait@uqu.edu.sa}
}

\maketitle  
\begin{abstract}
Multiple choice questions (MCQs) that can be generated from a domain ontology can significantly reduce human effort \& time required for authoring \& administering assessments in an e-Learning environment.  Even though there are various methods for generating MCQs from ontologies, methods for determining the difficulty-levels of such MCQs are less explored. In this paper, we study various aspects and factors that are involved in determining the difficulty-score of an MCQ, and propose an ontology-based model for the prediction. This model characterizes the difficulty values associated  with the stem and choice set of the MCQs,  and describes a measure which combines both the scores. Furthermore, the notion of assigning difficultly-scores based on the skill level of the test taker is utilized for predicating difficulty-score of a stem. We studied the effectiveness of the predicted difficulty-scores with the help of a psychometric model from the Item Response Theory, by involving real-students and domain experts. Our results show that, the predicated difficulty-levels of the MCQs are having  high correlation with their actual difficulty-levels.

\end{abstract}

\section{Introduction}

Finding the difficulty-level of computer-generated questions, has recently gained much attention in educational community as well as in computer science community. This is mainly due to the wide-spread use of computer based tutoring systems in guiding and assessing learners. When it comes to assessing the knowledge level of a learner using an evaluation test --- preliminary and concluding evaluations --- the difficultly-levels of the questions used will play a crucial role in classifying the learners based on their knowledge proficiencies~\cite{sage}.

An assessment is mainly conducted to determine the amount and quality of knowledge a learner has gained from a session (or a series of sessions) of knowledge transfer. Therefore, one of the desired properties of an assessment test is to discriminate between good, average and low level learners~\cite{sage3,sage2}. Reducing burden in authoring and administering a test is another desired property of an assessment. Usually for large-scale assessments, multiple choice questions (MCQs) are of great usefulness, as the test administration and evaluation process can be automated using a computer program.

In the literature, there have been many attempts such as~\cite{Cubric2010,EV2015,Heilman,flairs15} for generating MCQs. These approaches have used structured or unstructured knowledge sources for generating MCQs. Recent works were focusing more on structured knowledge sources like Web Ontology Language (OWL) ontologies, than unstructured knowledge sources (e.g., text file), for many reasons. Some of the reasons for this trend that are suggested by Alsubait et al.~\cite{simieee} are the richness of ontologies and their resemblance to knowledge representations acquired by learners in the course of their studies. In addition, OWL ontologies are based on a decidable fragment of first order logic, called Description Logics~\cite{Baader}, where reasoning can be done to infer implicit relations among the domain entities. This inferencing process is similar to the reasoning carried out by a learner to reach conclusions on a subject matter.

Even though there are numerous methods for generating MCQs, methods to estimate difficulty-levels of such MCQs are less explored. Without a proper difficulty-level estimation method, the generated MCQs are less employable in pedagogy applications.

In this paper, we propose an automated method to determine the difficulty-levels of MCQs that are automatically generated from ontologies. Furthermore, we discuss how the knowledge proficiency of a learner affects the difficulty-level of a question. Section~\ref{sec4} describes a method to check the sanity of the predicted difficulty values of the MCQs. In Section~\ref{secEE}, we describe an experiment which is based on a psychometric model, to determine the actual difficulty-levels of a set of machine-generated MCQs. We then correlate this actual difficulty-levels with their estimated (or predicated) difficulty-levels, to examine the effectiveness of our approach.

\section{Background and Motivation}
\subsection{Ontology-based MCQ generation systems}
Ontology-based MCQ generation was first introduced by Papasalouros et al.~\cite{Papas}, where they have introduced a set of strategies based on different ontological entities for framing MCQs and the corresponding distracting answers. Their method lacks proper theoretical support for when to use which strategy, and the stem of all the generated questions remains the same (\emph{\enquote{Choose the correct sentence}}). Later, Tosic and Cubric \cite{Cubric2010} generated MCQs of knowledge level (\emph{\enquote{Which of the following definition describes the concept $C$?}}), comprehension level (\emph{\lq Which one of the following response pairs relates in the same way as $a$ and $b$ in the relation $R$?}\rq), application level (\emph{\enquote{Which one of the following examples demonstrates the concept $C$?}}) and analysis level (\emph{\enquote{Analyze the text $x$ and decide which one of the following words is a correct replacement for the blank space in $x$.}}). They further extended the work introduced by Holohan et al.~\cite{extend}, by introducing stems that use annotation information in the ontology. Strategies similar to Papasalouros's strategies were adopted to find the distractors for the generated question statements. Another MCQ generation approach was by Alsubait et al.~\cite{mining}. Their approach was limited to analogy type questions. Later, they extended their work to include different kinds of MCQs \cite{lesson,Alsubait2015}. Other than the above MCQ generation approaches, there are works such as~\cite{Ben,zito}, which utilize simple ontology statements: concept inclusions, role hierarchy and (concept and role) assertions, to generate basic domain related questions.

Most recent works on MCQ generation can be found at~\cite{EV2015,flairs15,Alsubait2015}. In~\cite{EV2015}, the authors make use of the semantics of the domain, represented in the form of TBox axioms along with ABox axioms, to frame interesting MCQs. Vinu and Kumar~\cite{flairs15} illustrated an effective and practically adoptable generic (i.e., domain independent) method to generate MCQs from the assertional facts of a given ontology. Their method, \emph{pattern-based MCQ generation}, utilizes different combinations of predicates (i.e., concept names and roles) associated with the instances of an ontology, for framing the stems. As at present, the pattern-based approach is the state-of-the-art technique and since we can generalize existing methods to this approach, we predict the difficulty values of the MCQs that are generated using that method.  A detailed explanation of the pattern-based MCQ generation method is given in the next subsection to set the context for the current work. 

\subsection{Pattern-based MCQ generation}
In the pattern-based MCQ generation, introduced in~\cite{flairs15}, components of an MCQ --- \emph{Stem} and \emph{Choice set} (i.e., key and distractors) --- are generated using simple SPARQL templates. In this approach, a stem can be considered as a set of \emph{conditions} which asks for a solution which is explicitly present in the ontology. The set of conditions in a stem is formed using different combinations of (unary or binary) predicates associated with an instance in an ontology. Example-\ref{eg1} is such an MCQ, which is framed from the following assertions that are associated with the (\emph{key}) instance \emph{birdman}. For ease of reading, all examples presented in this paper are from the Movie domain.

{
\emph{Movie(birdman)}  //concept assertion

\emph{isDirectedBy(birdman,alejandro)}  //role assertion

\emph{hasReleaseDate(birdman,"Aug 27 2014")}  //role assertion
}

\begin{example}\normalfont \label{eg1}
{\bf Stem:} Choose a \emph{Movie}, which \emph{is directed by Alejandro} and \emph{has release date {Aug 27, 2014}}.
\begin{center}
\noindent
\begin{tabular}{ @{}lllll@{}}
 \midrule
 \multicolumn{2}{l}{{\bf Choice set:}}&~~~~~a. Birdman  &~~~~~b.  Titanic&~~~~~c.  Argo \\ 
 \midrule
\end{tabular}
\end{center}
\end{example}

The possible predicate combinations (or question patterns) of size\footnote{Signifies the number of predicates in a combination} one w.r.t. an instance $x$ can be denoted as: $x\,\vec{O}\, i,$ $x \,\cev{O}\, i\,,~x\, \vec{D}\, v$ and $x\, \vec{a}\, C$, where $i$ is an instance, $\vec{a}$ is \emph{ rdf:type}, $\vec{O}$ and $\cev{O}$ represents object properties that are having different directions, $\vec{D}$ denotes datatype property, $v$ stands for the value of the datatype property and $C$ is a class name. We call the instance $x$ as the \emph{pivot-instance} of the question pattern. The arrows ($\leftarrow$ and $\rightarrow$) represent the directions of the properties w.r.t. the pivot-instance. In~\cite{flairs15}, the authors studied the pattern combinations of sizes one and two; but, in this paper we are generalizing the pattern size to $n$ (a positive integer). A matching tuple of the pattern (of size two): $i_{1}\,\vec{O_{1}}\, x\, \cev{O_{2}}\, i_{2}$, is represented as a list: $[(a,O_{1},b), (c,O_{2},b)]$, where $b$ corresponds to the pivot-instance $x$. Linguistic representations of tuple lists are addressed as the \emph{stems}. For a pattern of size $n$, there will be $n$ ordered triples in the list. These triples without the pivot-instance is considered as the conditions of a stem. The conditions of a stem $S$ is represented as $C_{S}=\{r_{1}, r_{2},...,r_{n}\}$, where, for an instance $y$, \enquote{$r_{i}.y$ is true} means that $y$ satisfies the condition $r_{i}$. In this paper, we consider the pivot-instance of a tuple, as the \emph{key} corresponding to the stem generated from that tuple. We do not consider tuples containing blank nodes for stem generation.

The distractors for these MCQs are selected from the set of instances (or literals) of the ontology which belong to the intersection classes of the domain or range of the predicates that are present in the stem. In our earlier work~\cite{flairs15}, this set was addressed as \emph{Potential-set}. For example, the potential-set corresponding to the pattern $i_{1}\,\vec{O_{1}}\, x\, \cev{O_{2}}\, i_{2}$, is $\emph{Range}(O_{1})\sqcap\emph{Range}(O_{2})$. Readers are referred to the original paper, for more details.

The \emph{choice set} of an MCQ can be considered as a set of instances from its potential-set along with the key, presented in some random order. In this work, we consider the cardinality of the choice set as three (including exactly one key), as it is the optimal number of options preferred for conducting an MCQ test by many researchers~\cite{Dehnad2014398,haladyna}. 

\subsection{Difficulty-level calculation\label{simil}}
A simple notion to find the difficulty-level of an ontology-generated MCQs was first introduced by Cubric and Tosic~\cite{Cubric2010}. Later, in~\cite{simieee}, Alsubait et al. extended the idea and proposed a similarity-based theory for controlling the difficulty of ontology generated MCQs.  In~\cite{mining}, they have applied the theory on analogy type MCQs. In~\cite{lesson}, the authors have experimentally verified their approach in a student-course setup.  The practical solution which they suggested to find out the difficulty-level of an MCQ is w.r.t. the degree of similarity of the distractors to the key. If the distractors are very similar to the key,  students may find it very difficult to answer the question, and hence it can be concluded that the MCQ is difficult. For instance, the following table shows the similarity of the distractors with the key, for a given stem (see Example-\ref{eg1} for stem details), where the difficulty due to distractors = ${(0.5+0.4)}/{2} = 0.45$.
\begin{table}[!h]
\centering
\noindent\scalebox{1}{
\begin{tabular}{ @{}lllll@{}}
 \midrule
 \multicolumn{2}{l}{{\bf Choice set:}}
&~~~~~a. Birdman (\emph{Key}) 
&~~~~~b.  Titanic
&~~~~~c.  Argo \\ 
\midrule
\multicolumn{2}{l}{{\bf Similarity with the key:}}
&\multicolumn{1}{r}{1.0}  
&\multicolumn{1}{r}{0.5}
&\multicolumn{1}{r}{0.4} \\ 
\midrule
\end{tabular}}
\end{table}

Similarity-based theory~\cite{simieee} was the only effort which is utilized in the literature~\cite{EV2015,flairs15,lesson},  to calculate or control the hardness of an ontology-based generated MCQ. The difficulty value determined by similarity-based theory considers only the similarity of the correct-option (\emph{key}) with the wrong-options (\emph{distractors}). That is, if the similarity is high, then it implies that the MCQ has a high difficulty-level and vice versa. 
%

In many a case, the stem (question statement) of an MCQ is also a deciding factor for the hardness of an MCQ. For instance, the predicate combination (discussed in the previous subsection) which is used to generate a stem can be chosen such that they can make the MCQ harder or easier to answer. We investigate this aspect in Section~\ref{sec3}. Difficulty-level calculation based on Similarity-based theory is detailed in Section~\ref{sec4}. A unified method, by combining the two methods is proposed in Section~\ref{sec5}. An empirical study in a classroom setup (see Section~\ref{secEE}) was done to evaluate the effectiveness of the proposed method.

\section{Difficulty-level of Stem\label{sec3}}
Consider the stems (underlined portions indicate the concept names and roles  used): 
\begin{itemize}
\item[]{\bf Stem-1: }\emph{Choose a \underline{Movie} which was \underline{directed by} Clint Eastwood}.  
\item[]{\bf Stem-2: }\emph{Choose an \underline{Oscar movie} which was \underline{directed by} Clint Eastwood}. 
\end{itemize}
In the empirical study that we have conducted as a part of our earlier work, we observed that,  the former question, for a given set of options (we will see the significance of the \emph{choice set} later), is not correctly answered by many people with good knowledge about the movie domain, when compared to the latter question (especially when the choice set does not contain any famous movie names). A rational explanation for this behavior is that, \emph{Explanation-1:} in the latter question, it is given that the movie is an Oscar Movie, which is a \emph{good hint} for an expert to correctly answer the question, whereas in the former case, she will find it difficult to choose the right answer, as the hint is very poor in identifying the key. It should be noted that, this explanation is given in reference to the perspective of an expert-level learner. 


On the other hand, if we look from a beginner-level learner's perspective, she 
may identify the hint; but may not be able to utilize it due to lack of expertise. For her, the difficulty of a question will increase if the stem contains specific details of the domain. If we revisit the previous stems, Stem-2 is likely to be more difficult than Stem-1; this cannot be established using Explanation-1. Therefore, it is necessary to come up with a new explanation for question difficulty which should go with the beginner-level perspective.  

 
A logical explanation in reference to a beginner-level learner can be, \emph{Explanation-2:} when compared to Stem-1, Stem-2 requires the additional knowledge of Oscar movies for correctly answering the question. Therefore, the latter question is more difficult for a beginner-level learner than the former one. But, here we are not claiming that Stem-1 is easy for beginner-level learners; rather, we are saying that Stem-2 is relatively harder to answer than Stem-1, fulfilling the following  desired aspect.

A \emph{desired aspect} of a difficulty-level of a question is that, if a given question is difficult for an expert-level learner then, it should also be difficult for a beginner-level learner.

A proposition that can be made out from the aforementioned explanations based on the perspective of a learners knowledge-level is that, \emph{difficulty level of an MCQ is relative to the knowledge proficiency of a test taker}. In other words, the difficulty-score (value) of an MCQ should be defined in terms of the knowledge proficiency of the test takers. This indeed makes it necessary to have multiple difficulty-scores for a given MCQ. In this paper, we limit the scope of our studies by categorizing the knowledge proficiencies (a.k.a. \emph{trait-levels}) of learners into three: \emph{beginner, average} and \emph{expert}. We then formulate  difficultly measures for each of the three cases.

\subsection{Difficulty measure for expert-level learner\label{DME}}
As we have mentioned above, the difficulty-score of an MCQ w.r.t. to an expert learner depends on how good the hints are  helping her to answer the MCQ correctly. By hints we mean the conditions present in the stem. The obvious method to measure the quality of a hint is by measuring how well the \emph{answer space} of the question is affected by the hint; in the current context, answer space refers to the number of instances that satisfy a given condition. That is, if the answer space of the question is greatly reduced because of a particular hint, then it is a \emph{good hint}. For example, in the stems that we saw earlier, \emph{directed by Clint Eastwood} is a {good hint} --- as the count of instances which satisfy this condition is low. 
Since both stems have the same (good) hint, we are not gaining any insight about their relative hardness, just by looking at these good hints alone. We have to look at the other  predicates (here, the unary predicates: \emph{Movie} and \emph{Oscar Movie}) as well. These predicates clearly distinguish the two stems (i.e., Stem-2 is easier to answer than Stem-1), by looking at the count of instances satisfying these predicates. This observation, made us to conclude that, \emph{answer spaces of all the predicates in a stem should be considered for finding its difficulty-score.}

Now, consider another stem, {\bf Stem-3:} \emph{Choose a Thriller Movie which was directed by Clint Eastwood}. If the answer spaces for the predicates: \emph{Oscar Movie} and \emph{Thriller Movie}, happened to be of same cardinality, then, both stems (Stem-2 and Stem-3) will have the same difficulty-scores. On the contrary, we observed that, Stem-2 is less difficult for an expert to answer correctly than Stem-3. A rationale for this behavior is that, the concept \emph{Oscar Movie} is a more popular concept than \emph{Thriller-Movie}. Therefore, we consider, \emph{the popularities of the predicates present in a stem as another factor which affects its difficulty.}

We  propose the difficulty-score of a stem $S$, which is having predicates (concepts or roles) $P=\{p_{1},p_{2},...,p_{n}\}$ (a.k.a \emph{Property set}), generated from an ontology $\mathcal{O}$, in terms of the answer spaces (denoted as $ASpace_{\mathcal{O}}(.)$), and popularities of these properties (represented as $PredicatePopularity_{\mathcal{O}}(.)$), as shown below:
\begin{equation}\label{dexpert}\small
D_{expert}(S,\mathcal{O})=\sum_{i=1}^{n} \frac{\log(1+|ASpace_{\mathcal{O}}(p_{i},S)|)}{PredicatePopularity_{\mathcal{O}}(p_{i},S)}
\end{equation}

\paragraph{Answer space.}The answer space of $p$ in $S$ (i.e., $ASpace_{\mathcal{O}}(p,S)$) is the set of instances which satisfy the predicate $p$ in $\mathcal{O}$.  If $p$ is a unary predicate (i.e., a concept), then the definition is straight forward. But, to find the answer space  of a binary predicate, we consider the directionality of the predicate w.r.t. the {pivot-instance} of the corresponding pattern and accordingly choose the subject or object along with the predicate (this is called the \emph{concept equivalent} of the binary predicate). For example, consider the tuple corresponding to Stem-3: [(\emph{mystic\_river, is\_a, Thriller\_movie}), (\emph{mystic\_river, isDirectedBy, clint\_eastwood})], we find the instances which belong to the concept \emph{Thriller\_movie} and those instances which satisfy the constraint (i.e., the concept equivalent): $\exists$\emph{isDirectedBy.} \emph{\{clint\_eastwood\}}, to find the answer space. For a tuple of the form: [(\emph{mystic\_river, is\_a, Thriller\_movie}), (\emph{clint\_eastwood, isDirectorOf, mystic\_river})], since the directionality of the predicate in the second triple is different from that in the tuple for Stem-3, we take the concept equivalent as: $\exists$\emph{Inv(isDirectorOf)}.\{\emph{clint\_eastwood\}} --- if \emph{Inv(.)}, inverse of the predicate is not present, we look for the pattern (using the SPARQL query fragment:) \enquote{\emph{clint\_eastwood ~ isDirectorOf~ ?x}}  in the rdf graph.


\paragraph{Predicate Popularity.}We calculate the popularity of a predicate $p$ in $S$ ( i.e., ${Predicate}$ ${Popularity_{\mathcal{O}}(p,S)}$) 
by finding  
the mean of the popularities of all the instances which satisfy $p$ in $\mathcal{O}$; the popularity of an instance $i$, satisfying $p$ is denoted as $Popularity_{\mathcal{O}}(p,i)$. Here, $p$ can be a unary predicate or a binary predicate. If $p$ is a binary predicate, its concept equivalent  will be considered. A widely used  measure for finding the popularity of a concept $c$ is based on the count of the instances of the other concepts that are related to the instances that satisfy $c$~\cite{pop}. We make use of this measure to find the popularity of a concept. Therefore, ${PredicatePopularity_{\mathcal{O}}(.)}$ of a unary predicate $p$ of a stem $S$ can be calculated as follows, where $n$ is the cardinality of the set ${ASpace{_\mathcal{O}}(p,S)}= \{i_{1}, i_{2}, ..., i_{n} \}$:
\begin{equation}\label{mp}\small
PredicatePopularity_\mathcal{O}(p,S) = {(1/n) \times\sum_{j=1}^{n} Popularity_\mathcal{O}(p,i_{j})}
\end{equation}
 
In Eq.~\ref{mp}, $Popularity_\mathcal{O}(p,i)$ gives the count of the instances that are related to $i$ by a relation $R$, such that they do not satisfy $p$. 
 That is, $Popularity_\mathcal{O}(p,i) = |\{j|\mathcal{O}\models p(i) \sqcap R(j,i) \land \mathcal{O}\not\models p(j)\}|.$

\subsection{Difficulty measure for beginner-level learner\label{DMB}}
For a  beginner-level learner, the difficulty-level depends on how detailed the question is. Intuitively, if the stem contains domain specific conditions, the probability of a learner for correctly answering the question will reduce. Therefore, we relate the depths of the concepts and roles that are used in the stem to the concept and role hierarchies of the ontology, to determine the stem difficulty.  
To achieve this, we introduce ${DepthRatio}$ for each predicate $p$ in an ontology. ${DepthRatio}$ is defined as:
\begin{equation}\label{DRequ}\small
DepthRatio_\mathcal{O}(p)= \frac{\text{Depth (or length) of $p$ from the root of the hierarchy}}{\text{Maximum  length of the path containing $p$}}
\end{equation}

For a stem $S$, generated from an ontology $\mathcal{O}$, with $x$ as its pivot-instance and $P$ as its property set, let $\mathcal{C}$ denote the set of concepts satisfied by $x$, and let $\mathcal{R}$ represents the set of roles such that either $x$ is present at their domain (subject) or range (object) position \big(i.e., $R \in \mathcal{R} \implies \mathcal{O}\models R(x,i) \lor R(i,x)$, where $i$ is an arbitrary instance in $\mathcal{O}$\big). For each $p\in P$, we find the largest subset in $\mathcal{C}$  (if $p$ is a concept) or we find the largest subset in $\mathcal{R}$ (if $p$ is a role), such that the elements in the subset can be related using the relation $\sqsubseteq$, and $p$ is an element in that subset. The cardinality of such a subset forms the denominator of Eq.~\ref{DRequ}, and the numerator is the position of the predicate $p$ from right (right represents the top concept or top role), when the elements in the subset are arranged using the  relation $\sqsubseteq$.

In addition to the $DepthRatio$, answer space of the predicates in a stem also affect the difficulty-score of an MCQ (recall Stem-1 and Stem-2 in Section~\ref{sec3}). That is,  if the cardinality of the answer space increases, there is a tendency that stem becomes more generic. For example, if we include the condition $\exists$\emph{isDirectedBy.\{tim\_miller\}} in the stem rather than $\exists$\emph{isDirectedBy.}\emph{\{clint\_eastwood\}}, the answer space will largely reduce (as Tim Miller is a young director who has directed only a few movies), resulting in an increase in the difficulty-score.   Therefore, we define the difficulty-score of a stem having $n$ number of predicates, in terms of the depth ratios and answer space of its predicates, as follows.
\begin{equation}\label{dbegin}\small
D_{beginner}(S,\mathcal{O})=\sum_{i=1}^{n} DepthRatio_\mathcal{O}(p_{i}) \times  \frac{1}{1+\log(1+|ASpace_\mathcal{O}(p_{i},S)|}
\end{equation}

The portion of the equation: $1/{(1+\log(1+|ASpace_\mathcal{O}(p_{i},S)|)})$, shows that the difficulty-score of a stem is inversely proportional to the cardinalities of the answer spaces of its predicates.  
Including a factor of depth ratio in Eq.~\ref{dbegin}, will help to make the difficulty-score high, when the stem contains more specific predicates.

\subsection{Difficulty measure for average-level learner\label{DMA}}
For a learner with average-level knowledge proficiency, both the factors which we discussed in Section~\ref{DME} and Section~\ref{DMB}, need to be considered for calculating the stem difficulty. Therefore, we take the mean of $D_{beginner}$ and $D_{expert}$, to find $D_{average}$.
\begin{equation}\small
D_{average}(S,\mathcal{O})=(1/2)\times\big({D_{expert}(S,\mathcal{O}) + D_{beginner}(S,\mathcal{O})}\big)
\end{equation}

\section{Validity of stems\label{sec4}}
As we have seen above, given an MCQ stem $S$, generated from an ontology $\mathcal{O}$, we can have three difficulty-scores (i.e., $D_{expert}(S,\mathcal{O}), D_{average}(S,\mathcal{O}), D_{beginner}(S,\mathcal{O})$)  assigned to it, corresponding to expert, average and beginner level learners, respectively. Let $d_{e}, d_{a}$ and $d_{b}$ be the normalized\footnote{Normalization is done by dividing the scores by the maximum score.}
 scores ($\rightarrow [0,1]$) of $D_{expert},D_{average}$ and  $D_{beginner},$ respectively.  Since $d_{a}$ is the average of $d_{e}$ and $d_{b}$, we can represent the difficulty-score of $S$ in terms of the tuple ($d_{e}, d_{b}$). 
 Based on the notion that if an expert finds a stem to be difficult then it is difficult for a beginner as well, and based on the desired property of a good assessment test (i.e., it should be able to distinguish expert, average and beginner level learners), we fix the validity definition of a given MCQ stem as:

\begin{definition}Validity of a stem. A given stem $S$,  with difficulty-score tuple ($d_{e}, d_{b}$), is \emph{valid}, if and only if  $d_{e} < d_{b}$.
\end{definition}

We ensure that the generated MCQs are valid as per the above definition before they are used in the experiment for empirical evaluation.

\section{Difficulty-level due to Choice set\label{sec4}}
\subsection{Similarity as a factor}

 In the similarity based method (see Section~\ref{simil}), the similarity measure for finding the instances' degree of similarity is subjective to the employed stem generation technique. For instance, in our earlier work~\cite{EV2015},  we proposed the \emph{Label-set similarity ratio} for finding the closeness of instances (where we associate TBox based restrictions that are present in the stem, to the instances to calculate their similarity); this measure cannot be used for those MCQs which are generated using pattern-based methods, since TBox axioms are not directly involved in the stem generation. 


Now the question is, how to calculate the similarity of two instances w.r.t. a stem $S$ from a given ontology $\mathcal{O}$.
Since there are no measures in the literature which satisfy our requirement, we propose a similarity measure, called \emph{Instance Similarity Ratio}, in Def.~\ref{def2}.

Given an ontology $\mathcal{O}$, two instances $k$ (the key) and $d$ (a distractor), and a stem $S$, to calculate the Instance Similarity Ratio, we consider (1) similarity of the instances w.r.t. the conditions (represented as $C_{S}=\{ r_{1}, r_{2},..,r_{n}\}$) in $S$. (The conditions in $S$, that are satisfied by $k$ and $d$, can be represented as $C_{k}=\{r|r.k \text{ is true in } \mathcal{O} \land r \in C_{S}\}$ and $C_{d}=\{r|r.d \text{ is true in } \mathcal{O} \land r \in C_{S}\}$ respectively, where $r.x$ is true if the condition $r$ is satisfied by the instance $x$.); (2) generic similarity of the two instances w.r.t. all the conditions they satisfy. (The set of all the conditions satisfied by an instance $i$ in $\mathcal{O}$ is represented as $A_{i}=\{ r|r.i~\text{is true in }\mathcal{O}\}.$) 

\begin{definition}\label{def2}Instance Similarity Ratio,  $Sim_{\mathcal{O}}(.),$ is defined as: 

\begin{equation}\label{sim}\small
Sim_{\mathcal{O}}(k,d)=\frac{\#\text{Instances in $\mathcal{O}$ that satisfy }C_{k}\cup C_{d}}{\#\text{Instances in $\mathcal{O}$ that satisfy }C_{k}\cap C_{d}} +  GSim_{\mathcal{O}}(k,d)^{2}
\end{equation}

The first part of the right hand side denotes the similarity of $k$ and $d$, w.r.t. the conditions in $S$. The second part, $GSim_{\mathcal{O}}(.)$, denotes the generic similarity of the instances.
\begin{equation}\small
GSim_{\mathcal{O}}(k,d)=\frac{1}{2}\times\frac{\#\text{Instances in $\mathcal{O}$ that satisfy }A_{k}+ \#\text{Instances in $\mathcal{O}$ that satisfy }A_{d}}{\#\text{Instances in $\mathcal{O}$ that satisfy}A_{k}\cap A_{d}} 
\nonumber
\end{equation}

$A_{k}$ and $A_{d}$ represent the set of all the conditions that are satisfied by $k$ and $d$ respectively in $\mathcal{O}$. 

\end{definition}

The first part of Eq.~\ref{sim} gives the ratio of the number of instances in $\mathcal{O}$ that satisfy all the conditions that are satisfied by $k$ and $d$, to the number of instances that satisfy the conditions that are satisfied by both $k$ and $d$. This formula is similar to the \emph{Jacquard similarity} measure. To ensure that the selected distractor $d$ has a minimum similarity with $k$, we consider $C_{k}\cap C_{d} \not = \phi$. The second part of the equation gives the square of the similarity of the two instances which is obtained by considering all the conditions they satisfy.

\begin{definition}Difficulty-score due to Choice set. 
The difficulty-score due to the choice set $X=\{ k,d_{1},d_{2},d_{3},...,d_{l}\}$ (with $k$ as the key) of an MCQ  is defined as:
\begin{equation}\small
D_{similarity}(X,\mathcal{O})= (1/l) \times{\sum_{i=1}^{l} Sim_{\mathcal{O}}(k,d_{i})}
\end{equation}
\end{definition}
\subsection{Popularity as a factor}

It is observed that the popularities of the instances that are being chosen as the key and as the distractors have a negative impact on the overall difficulty-score of the MCQ. For instance, consider the choice sets in Example~\ref{eg21}. The key and the distractors in the first choice set have relatively high popularities (due to more number of connectivity) than those in the second choice set. Intuitively, it is more likely that the most popular options are easily be identified by a learner as a correct answer or as a wrong answer. To capture this notion, we came up with an equation to find the difficulty-score due to popularity of a choice set as follows:

\begin{definition}{Difficulty-score due to popularity.} The difficulty-score due to the popularities of the instances in a choice set
$X =\{k,d_{1},d_{2},...,d_{l}\}$ (generated from an ontology $\mathcal{O}$) where $k$ is the key and $\{ d_{1},d_{2},...,d_{l} \}$ are the distractors, is defined as:
\begin{equation}\label{eqDP}\small
D_{popularity}(X,\mathcal{O}) = \frac{1}{Popularity(X,\mathcal{O})}
\end{equation}
The popularity of an instance $x$ in $X$ can be said as the {\bf connectivity} of $x$ from the other individuals in $\mathcal{O}$ which belong to a concept which $x$ does not belong to.  $\mathcal{C}_{\mathcal{O}}(j)$ represents the connectivity (formally defined below) of an instance $j$ in $\mathcal{O}$. 
\begin{equation}\label{eqDP2}\small
 Popularity(X,\mathcal{O}) =
(1/2) \times Popularity_{\mathcal{O}}(k)+ \sum_{i=1}^l \log \big(1+\mathcal{C}_{t,\mathcal{O}}(d_{i})\big)
\end{equation}
\end{definition}


In Eq.~\ref{eqDP2}, the popularity is calculated by finding the sum of the { connectivities} of its instances, by giving more preference to the connectivity value of the key instance. 
--- i.e., the sum of half the connectivity value of the $k$ and $\log$ of the connectivity values of the distractors. This is because, given a popular instance as a key and less popular instances as distractors, a learner can easily figure out the correct answer, where as given a less popular instance as key and including a popular instance as distractor,  still can deviate the learner from answering the MCQ correctly. For this reason, while calculating the difficulty due to distractors, we divide it with  the overall popularities of the instances in the choice set, such that if the popularity of the choice set increases the MCQ will become more easier to answer.


\begin{definition}Connectivity.
The connectivity of an instance $x$ in ontology $\mathcal{O}$ is defined as follows, where $C_{x}$ and $C_{y}$ are concepts in  $\mathcal{O}.$
\[{\begin{array}{c}\small
\mathcal{C}_{t,\mathcal{O}}(x)=\#\{y ~|~ \mathcal{O} \models C_{x}(x) \sqcap C_{y}(y) \sqcap R(y,x) \land \mathcal{O} \not\models C_{y}(x) \land\\ not(C_{x} \sqsubseteq C_{y}) \land not(C_{y} \sqsubset C_{x}) \}
\end{array}}\]
\end{definition}
The equation gives the count of the instances which are related to $x$ by  a relation $R$ and whose satisfying concepts do not satisfy $x$ and  are not hierarchically (sub-class--super-class relationship) related to any of the satisfying concepts of $x$.

\begin{example}\normalfont \label{eg21}
{\bf Stem:} Choose an \emph{American martial arts film.}\\
\begin{center}
\small
\begin{tabular}{ @{}lllll@{}}
 \midrule
 \multicolumn{2}{l}{{\bf Choice set 1:}}
&~~~a. The Karate Kid (\emph{Key})  
&~~~~b.  102 Dalmatians
&~~~~~c.  Gone with the Wind \\ 
\midrule
\multicolumn{2}{l}{{\bf Popularity:}}
&\multicolumn{1}{r}{11}  
&\multicolumn{1}{r}{8}
&\multicolumn{1}{r}{10} \\ 
\midrule
\end{tabular}\\
\begin{tabular}{ @{}lllll@{}}
 \midrule
 \multicolumn{2}{l}{{\bf Choice set 2:}}
&~~~a. Sunland Heat (\emph{Key})  
&~~~b. A Vanished World
&~~~c. Just Call Me Nobody \\ 
\midrule
\multicolumn{2}{l}{{\bf Popularity:}}
&\multicolumn{1}{r}{4}  
&\multicolumn{1}{r}{3}
&\multicolumn{1}{r}{2} \\ 
\midrule
\end{tabular}
\end{center}
\end{example}

\subsection{Overall difficulty-score due to choice set}

The overall difficulty-score due to the choice set (denoted as $DC$) of an MCQ generated from an ontology $\mathcal{O}$ with stem $S$ and choice set $X$ is defined as:
\begin{equation}\small
DC(X,\mathcal{O})= D_{similarity}(X,\mathcal{O}) \times D_{popularity}(X,\mathcal{O})
\end{equation}

\section{Combined Score: Difficulty-level of MCQ\label{sec5}}
Having the difficulty-score of a stem $S$ (denoted as $DS$) and the difficulty-score due to its choice set $X$ (represented as $DC$), we can find the overall difficulty-score of the MCQ $Q=(S,X)$ (generated from ontology $\mathcal{O}$) by simply multiplying $DS$ and $DC$. We choose to multiple the two scores, since, either a poor set of options or a easy stem can make an MCQ less difficult to answer.
\begin{equation}\label{eqoverall}\small
D_{predicted}(Q) = DS(S,\mathcal{O}) \times DC(X,\mathcal{O})
\end{equation}

In the Eq.~\ref{eqoverall}, depending on the pedagogical goal of the assessment (for example, whether the test MCQs are meant for identifying expert learners) we will choose $D_{beginner}, D_{average}$ or $D_{expert}$, as $DS$.

\section{Empirical Evaluation}\label{secEE}

We have implemented a prototype system  (developed using Java Runtime Environment JRE V1.6 and the Jena API\footnote{https://jena.apache.org/} 2.11.0) which can generate MCQs and predict their difficulty-scores. The system is being tested by giving various ontologies (mentioned in~\cite{EV2015,flairs15}) as inputs. 
In this section, we mainly investigate the effectiveness of the proposed difficulty-score predication measures. 
For this purpose, we have generated  test MCQs from a handcrafted ontology, and  determined their difficulty-scores  (a.k.a \emph{predicted difficulty-scores}), using the method proposed in Section~\ref{sec5}. Then, we compared these predicted difficulty-levels with their \emph{actual difficulty-levels} that were estimated using principles in the Item Response Theory. 


\subsection{Estimation of actual difficulty-level}
Item Response Theory is an item oriented theory which specifies the relationship between learners' performance on test items and their ability which is measured by those items~\cite{sage}. 
In IRT, \emph{item analysis} is a popular procedure  which tells if an MCQ is too easy or too hard, and how well it discriminates students of different knowledge proficiencies. Here, we have used item analysis to find the actual difficulty-levels of the MCQs. 

We make use of the simplest IRT model (often called \emph{Rasch model} or the \emph{one-parameter logistic model} (1PL)) for finding the actual difficulty-level of the MCQs. According to this model, we can predict the probability ($P$) of answering a particular item (with difficulty value $\alpha$) correctly by a learner of  knowledge proficiency level $\theta$ (a.k.a \emph{trait level}), as  specified in the following formula.
\begin{equation}\label{hdnsX}\small
\begin{split}
\emph{P}(\theta,\alpha) = \frac{e^{(\theta - \alpha)}}{1+e^{( \theta - \alpha)}}
\end{split}
\end{equation}

Due to page limitation, we are providing the detailed theoretical background of the 1PL model in an online appendix\footnote{https://www.overleaf.com/read/vygnvzgcjvrg (Date of access:28/04/2016)}. To find the (actual) difficulty value, we can rewrite  Eq.\ref{hdnsX} as follows:
\begin{equation}\label{vvX}\small
\begin{split}
\alpha  = \theta -  log_{e}\Big(\frac{\emph{P}(\theta,\alpha)}{1-\emph{P}(\theta,\alpha)}\Big)
\end{split}
\end{equation}

For experimental purposes, suitable $\theta$ values can be assigned for high, medium and low trait levels. Given the probability (of answering an MCQ correctly by learners) of a particular trait level,  if the calculated $\alpha$ value is (approximately) equal or greater than the $\theta$ value, we can assign the trait level as its \emph{actual difficulty-level}.

\subsection{Experimental setup}
A controlled set of MCQs from the DSA ontology has been used to obtain evaluation data related to its quality. 
\paragraph{DSA ontology.} Data-structures $\&$ Algorithms (DSA) ontology (available at our project website\footnote{https://sites.google.com/site/ontomcqs/item-difficulty-results  (Date of access:28/04/2016)}) models the various aspects of data-structures and algorithms. An initial version of this ontology was developed by Revuri and Kumar~\cite{RevuriUK06} for information retrieval application. Later, an enriched version of the ontology which is primely focusing on question generation application is prepared by our research team. The ontology is composed of 90 concepts, 30 object properties, 13 datatype properties and 165 individuals.

\paragraph{Question-set generation:} For the experiment mentioned in this paper, we considered only three patterns (Pattern-1: $i\,\vec{O}\, x\, \vec{D}\, v$; Pattern-2: $i_{1}\,\vec{O_{1}}\, x\, \vec{O_{1}}\, i_{2}$ ; Pattern-3: $x\, \vec{O}\, i$ --- as they are the common patterns) for generating MCQ stems from the DSA ontology. Using  Pattern-1, we have generated $38$ stems, only $16$ stems were determined as valid stems. $368$ stems were generated using Pattern-2, only $169$ stems were predicated as valid stems. For Pattern-3, out of $405$ generated stems, $185$ were valid. These valid stems were then associated with choice sets such that their overall difficultly-level fall into the category of high, medium or low. 

\paragraph{Test MCQs and instructions.} 
We have administered a question-set of 24 (valid) test MCQs  to 54 participants, with the help of a web interface. 
These 24 MCQs were chosen such that, the test contains 8 stems of each (predicated) difficulty-level (high, medium and low). The difficulty-scores of their stems were pre-determined using the method detailed in Section~\ref{sec5}.  Difficulty-levels (predicted difficulty-levels) were then assigned by statistically finding  
suitable range of values (corresponding to low, medium and high) from the obtained normalized difficulty-scores of all the stems. Suitable choice sets were then assigned to further tune their difficulty-levels. More details about the question-set preparation and the question-set itself are available at our project website. All the test MCQs were  carefully vetted by human-editors to correct grammatical and punctuation errors, and to capitalize the proper nouns in question stems. Each MCQ contains choice set of cardinality three (with exactly one key) and two additional options: SKIP and INVALID. A sample MCQ is shown in Example-\ref{eg2}.
\begin{example}\normalfont \label{eg2}
Choose an Internal Sorting Algorithm with worse case time complexity\\ n exp 2.\\

\centering
\noindent
\begin{tabular}{ @{}lllll@{}}
 \midrule
 \multicolumn{2}{l}{{\bf Choice set:}}&~~~~~a. Heap Sort  &~~~~~b.  Bubble Sort&~~~~~c.  Breadth First Search \\ 
 
&&~~~~~d.  SKIP&~~~~~e. INVALID\\\midrule
\end{tabular}
\end{example}

The responses from  (carefully chosen) 54 participants --- 18 participants of each trait level (high, medium and low) --- were considered for generating the statistics about the item quality. The following instructions were given to the participants before starting the test. 

\begin{enumerate}
\item The test should be finished in 40 minutes and all questions are mandatory. 
\item You may tick the option \enquote{SKIP} if you are not sure about the answer. Kindly avoid guess work. 
\item If you find a question invalid, you may mark the option \enquote{INVALID}.
\item Avoid use of the web or other resources for finding the answers.
\item In the end of the test, you are requested to enter your expert level in the subject w.r.t this test questions, in a scale of high, medium or low. Also, kindly enter your grade which you received for the ADSA course offered by the Institute. 
\end{enumerate}


\paragraph{Participant selection.} Fifty four learners of the required knowledge proficiencies were selected from a large number of graduate level students (of IIT Madras\footnote{https://www.iitm.ac.in/}), who have participated in the online MCQ test. To determine their trait levels,  we have instructed them to self assess their knowledge-confidence level on a scale of high, medium or low, at the end of the test. To avoid the possible errors that may occur during the self assessment of  trait levels, participants with high and medium trait levels were selected from only those students who have successfully finished the course: CS5800: \emph{Advanced Data Structures and Algorithms}, offered at the computer science department of IIT Madras. The participants with high trait level were selected from those students with either of the first two grade points (i.e., 10 - Excellent and 9 - Very Good). The participants with medium trait level were from those students who were having any of the next two grade points (i.e., 8 - Good and 7 - Satisfactory Work). The evaluation data collected for the item analysis is shown in Table~\ref{t10a}.

\subsubsection{Item analysis}
The probabilities of correctly answering the test MCQs  (represented as $P$) by the learners are listed in Table~\ref{t10a}. In the table, the learner sets $L_{1}, L_{2}$ and $L_{3}$ correspond to the learners $l_{1}$ to $l_{18}$, $l_{19}$ to $l_{36}$ and $l_{37}$ to $l_{54},$ respectively. The learners in these learner sets have  high, medium and low trait levels, respectively. The probability $P$ (a.k.a., $P_{qr}$) of correctly answering  an MCQ $q$ for each of the learner sets ($L_{r}$) are obtained by dividing the count of learners in $L_{r}$ who have correctly answered $q$ by the cardinality of $L_{r}$. 
While calculating the $P$ values, if a learner has chosen the option \enquote{SKIP} as the answer, the MCQ is considered as wrongly answered by her. If she has chosen \enquote{INVALID}, we do not consider her poll for calculating $P$. 

In Table~\ref{t10a}, we list the $\alpha_{i}$ (actual difficulty) values that we have calculated using the corresponding $P$ values. Eq.\ref{vvX} is used for finding the \emph{difficulty} values. These $\alpha_{i}$ values were then used to assign the actual {difficulty}-level for the MCQs.
 \begin{table}[!htb]\caption{Thumb rules for assigning difficulty-level, based on IRT\label{t66}}\centering
\begin{tabular}{l c l}
\toprule
Trait level&$\alpha_{i}$&Difficulty-level\\\midrule
High &$(> 1.5)$ or $(\approx 1.5 \pm .45)$&High\\
Medium &$(> 0)$ or $(\approx 0 \pm .45)$&Medium\\
Low &$(> -1.5)$ or $(\approx -1.5 \pm .45)$&Low\\\bottomrule
\end{tabular}
\end{table}

 We are particularly interested in the highlighted rows in Table~\ref{t10a}, where 
an MCQ can be assigned an actual difficulty-level as shown in Table~\ref{t66}. That is, for instance, if the trait level of a learner is high and  $\alpha_{i}$ is  approximately equal to $\theta_{l}$ (ideally, $\alpha_{i} \geq \theta_{l}$), then, according to the IRT model, a difficulty-level of \enquote{high} can be assigned. In our experiments, to calculate $\alpha_{i}$ values for high, medium and low trait levels, we used $\theta_{l}$ values $1.5,0$ and $-1.5$ respectively.

\subsubsection{Results}

Out of the eight test MCQs which were having high predicted difficulty-levels, seven MCQs have high actual difficulty-levels. Among the eight medium difficulty-level MCQs, only one MCQ has deviated from its predicted difficulty-level.  Three out of eight MCQs having low predicted difficulty-levels showed no correlation with their actual difficulty-levels. The overall correlation of the predicated difficulty-levels of the MCQs with their actual difficulty-levels is $79$\%.

\subsubsection{Discussion and future work}

Even though the results of our difficulty-level predication method have shown a high correlation with the actual difficulty-levels, there are cases where the approach had failed to give a correct predication. 
In our observation, the repetition of similar words or part of a phrase in an MCQ's stem and its key,  is one of the main reasons for this unexpected behavior. This word repetition can give a hint to  the learner, enabling her to choose the correct answer. Example-\ref{eg3} shows the MCQ item $i_{15}$, where the repetition  of the word \enquote{string} in the stem and in the key, and appearance of the stem component \enquote{Robin-Karp Algorithm} in the choice set has degraded the MCQ's (actual) difficulty-level. Since our approach is not designed to identify such flaws in the questions, the predicted characteristics of the MCQs may deviate from their actual characteristics.

\begin{example}\normalfont \label{eg3}
{\bf Stem:} Choose a \emph{String Matching Algorithm}, which \emph{is faster than Robin-Karp Algorithm}.
\begin{center}
\noindent
\begin{tabular}{ @{}lllll@{}}
 \midrule
 \multicolumn{2}{l}{{\bf Choice set:}}&~~a. Robin-Karp algo.   &~~b.  Bubble sort&~~c.  Boyer Moore string search algo. \\ 
 \midrule
\end{tabular}
\end{center}
\end{example}

A validity check based on the quality assurance guidelines of an MCQ question (suggested by Haladyna et al. in~\cite{haladyna}) has to be done prior to finding the difficulty-levels of the MCQs. This would prevent the MCQs that have the the above mentioned flaws becoming part of the assessment test.

Furthermore, we observed that, out of $811$ stems that are generated for the experiment, only $46$\% of the stems were found to be valid. An initial analysis of these invalid stems has shown that some of these stems are wrongly marked as invalid; showing  \emph{false negative} errors. A detailed analysis of these false negatives has to be done in future, to further  refine the difficulty measures.

 Another future work is based on our observation that, Eq.~\ref{dexpert} for $D_{expert}$  cannot correctly capture the  difficulty-scores for all the cases. For example, when the $ASpace$ is very small, the equation tends to give a small difficulty-score, but (intuitively) in reality this may not be the case. When the question pattern becomes very rare then it will instead become difficulty for an expert to answer it correctly. Similar is the case when the $ASpace$ is too large, the equation gives a high difficulty-score, but in reality this may not be the case, as when the $ASpace$ becomes too large, the question becomes too generic that an expert will easily answer it. We came across similar observations for the equation which we used to calculate $D_{beginner}$. Due to the page limit, we refrain from giving more details. A detail empirical study is required to validate our observations.

\section{Conclusion}
We have proposed an item difficulty model for MCQs that are generated from formal ontologies. Various factors such as popularities of the ontology entities, answer space (number of instances in a class) and hierarchical depth of predicates, were considered as a part of the difficulty model for predicting MCQs' difficulty values. A criteria for checking the validity of the predicated difficulty-scores is also detailed. A detailed empirical study under real world conditions (making use of the Item Response Theory principles) is conducted to find out the actual difficulty-levels of the automatically generated MCQs; a comparison of the predicated difficulty-levels with their actual difficulty-levels has shown a great correlation. 

In this paper we have tried to characterize some of the factors that are associated with the difficulty of an MCQ question. Our experiments cannot conclusively show that the characterization of the factors is absolute or the factor which have been considered are the only factors which determine the difficulty-level of an MCQ. However, our empirical evaluation has shown that the measures which we proposed can correctly predict the difficulty-score of an ontology generated MCQ to a large extent.



\begin{table}[H]
\centering
\caption{The probabilities of correctly answering  the test MCQs (i.e., $P$ values) and the $\alpha_{i}$ values calculated using the obtained $P$ values are shown below. The learners in $L_{1}, L_{2}$ and $L_{3}$ are having high, medium and low domain knowledge proficiencies respectively. 
\label{t10a}}
\scalebox{0.9}{
\begin{tabular}{@{}cr@{~}r@{~}r@{~~~~~~~~} r@{~}r@{~}r@{~~~}cc@{ }}
\toprule
\multirow{3}{*}{MCQ item No.}&
\multicolumn{3}{c}{$P$ values for}&
\multicolumn{3}{c}{$\alpha_{i}$ values for }&
\multicolumn{1}{c}{\multirow{2}{*}{Actual}}&
\multicolumn{1}{c}{\multirow{2}{*}{Predicted}}\\

&\multicolumn{3}{c}{the learner sets}&
 \multicolumn{3}{c}{the learner sets}&
 \multirow{2}{*}{Diffi.-level}&\multirow{2}{*}{Diffi.-level}\\

&$L_{1}$&$L_{2}$&$L_{3}$&$L_{1}$&$L_{2}$&$L_{3}$&&\\\midrule

        $i_{1}$ &   $~~0.32$   &  $~~0.12$     &  $~~0.08$ & 
\cellcolor[gray]{0.8}{$2.25$}  &  $1.99$       &  $0.94$  &$high$&$high$   \\

          $i_{2}$ &   $0.43$   & $~~0.32$    & $~~0.12$ & 
\cellcolor[gray]{0.8}{$~~~~~1.78$}  & $~~~~~0.75$      & $~~~~~0.49$     &$high$&$high$   \\

          $i_{3}$ &   $0.45$   & $~~0.22$     &  $~~0.00$ & 
\cellcolor[gray]{0.8}{$1.70$}  & $1.27$       & $+\infty$ &$high$&$high$   \\

          $i_{4}$ &   $0.51$   & $~~0.42$    & $~~0.06$   & 
\cellcolor[gray]{0.8}{$1.46$}  & $0.32$      & $1.25$  &$high$&$high$   \\

          $i_{5}$ &   $0.58$   & $~~0.27$     & $~~0.00$  & 
\cellcolor[gray]{0.8}{$1.18$}  & $0.99$       & $+\infty$ &$high$&$high$   \\

          $i_{6}$ &   $0.44$   &  $~~0.30$     &  $~~0.11$  & 
\cellcolor[gray]{0.8}{$1.74$}  & $0.85$      & $0.59$ &$high$&$high$   \\

          $i_{7}$ &   $0.41$    &  $~~0.17$     &  $~~0.08$   & 
\cellcolor[gray]{0.8}{$1.86$} & $1.59$      & $0.94$  &$high$&$high$   \\

           $i_{8}$ &   $0.72$    &  $~~0.41$     &  $~~0.06$    &  
\cellcolor[gray] {0.8}{$0.56$}   & {$0.36$} & $1.25$   &$--$&$high$   \\\midrule

$i_{9}$   & $~~~0.96$ &                   $~~~0.51$   &  $~~~0.02$  &  
            $-1.68$ & \cellcolor[gray]{0.8}{$-0.04$}  & $2.39$&$medium$&$medium$\\

$i_{10}$ &   $0.94$    &                       $0.22$   &  $0.06$   &  
             $-1.25$ & \cellcolor[gray]{0.8}{$1.24$} & $1.25$&$medium$&$medium$\\

$i_{11}$ &   $1.00$    &                       $0.46$   &  $0.00$  &  
             $-\infty$   & \cellcolor[gray]{0.8}{$0.16$} & $+\infty$&$medium$&$medium$\\

$i_{12}$ &  $0.90$                          &  $0.56$    &  $0.11$  & 
            $-0.70$    & \cellcolor[gray]{0.8}{$-0.24$} & $0.59$&$medium$&$medium$\\

$i_{13}$ &  $1.00$     &                       $0.61$   & $0.08$ & 
            $-\infty$  & \cellcolor[gray]{0.8}{$-0.45$} & $0.94$&$medium$&$medium$\\

$i_{14}$ &  $1.00$     &                       $0.44$   &  $0.00$  &  
            $-\infty$    & \cellcolor[gray]{0.8}{$0.24$} & $+\infty$&$medium$&$medium$\\

$i_{15}$ &  $1.00$     &                       $0.89$   &  $0.40$  &  
             $-\infty$  & \cellcolor[gray]{0.8}{$-2.09$}  & $-1.09$&$--$&$medium$\\

$i_{16}$ &  $1.00$     &                       $0.48$   &  $0.06$   &  
            $-\infty$  & \cellcolor[gray]{0.8}{$0.08$}  & $1.25$ &$medium$&$medium$\\\midrule

$i_{17}$ &  $1.00$     &  $0.82$                        &  $0.38$    &  
            $-\infty$  &  $-1.52$  & \cellcolor[gray]{0.8}{$-0.84$}&$--$&$low$\\

$i_{18}$ &  $1.00$     &  $0.84$                          &  $0.16$  &   
            $-\infty$  &  $-1.66$    & \cellcolor[gray]{0.8}{$0.16$}&$--$&$low$\\

$i_{19}$ &  $1.00$     &  $1.00$                        &  $0.48$   &   
            $-\infty$ &   $-\infty$& \cellcolor[gray]{0.8}{$-1.42$}&$low$&$low$\\

$i_{20}$ &  $0.96$    &  $0.88$                       &  $0.50$  &   
            $-1.68$ &  $-1.99$  & \cellcolor[gray]{0.8}{$-1.50$}&$low$&$low$\\

$i_{21}$ &  $1.00$     &  $0.92$                         &  $0.32$   &  
            $-\infty$  &  $-2.44$  & \cellcolor[gray]{0.8}{$-0.75$}&$--$&$low$\\

$i_{22}$ &  $1.00$     &  $0.96$                         &  $0.26$ &  
            $-\infty$  &  $-3.18$   & \cellcolor[gray]{0.8}{$-1.05$}&$low$&$low$\\

$i_{23}$ &  $1.00$     &  $0.77$                       &  $0.22$   &   
            $-\infty$    &  $-1.21$  & \cellcolor[gray]{0.8}{$-1.18$}&$low$&$low$\\

$i_{24}$ &  $1.00$     &  $0.88$                          &  $0.00$  &
            $-\infty$  &  $-1.99$    & \cellcolor[gray]{0.8}{$-1.26$}&$low$&$low$\\
\bottomrule
\end{tabular}}
\end{table}

\bibliography{ref.bib}{}
\bibliographystyle{splncs}

\end{document}